\documentclass[12pt]{article}

\usepackage{scicite}

\usepackage{times}

\newenvironment{sciabstract}{%
\begin{quote} \bf}
{\end{quote}}

\newcounter{lastnote}

\usepackage[utf8]{inputenc} 
\usepackage[T1]{fontenc}    
\usepackage{hyperref}       
\usepackage{url}            
\usepackage{booktabs}       
\usepackage{amsfonts}       
\usepackage{nicefrac}       
\usepackage{microtype}      
\usepackage{xcolor}         
\usepackage{amsmath}
\usepackage{amssymb}

\usepackage{microtype}
\usepackage{graphicx}
\usepackage{subcaption}
\usepackage{booktabs} 
\usepackage{multirow}
\usepackage{amsmath}
\usepackage{amssymb}
\usepackage{wrapfig}
\usepackage{lipsum}

\newcommand\blfootnote[1]{%
  \begingroup
  \renewcommand\thefootnote{}\footnote{#1}%
  \addtocounter{footnote}{-1}%
  \endgroup
}

\usepackage{hyperref}

\usepackage{cleveref}
\usepackage{mathtools}
\usepackage{xparse}
\usepackage{xcolor}
\usepackage{amsthm}
\usepackage{amsmath}
\newtheorem{definition}{Definition}
\newtheorem{principle}{Principle}
\newtheorem{theorem}{Theorem}

\theoremstyle{definition}

\title{A Theory of Human-Like Few-Shot Learning}

\author
{
Zhiying Jiang$^{1}$,  Rui Wang$^{2}$, Dongbo Bu$^{2}$, Ming Li$^{1\ast}$ \\
\normalsize{$^{1}$David Cheriton School of Computer Science, University of Waterloo,}\\
\normalsize{200 University Ave W, Waterloo, ON N2L 3G1, Canada}\\
\normalsize{$^{2}$Institute of Computing Technology, Chinese Academy of Science, Beijing, China}\\
\normalsize{$^\ast$To whom correspondence should be addressed; E-mail: mli@uwaterloo.ca}
}

\date{}

\begin{document}

\maketitle

\begin{sciabstract}
We aim to bridge the gap between our common-sense few-sample human learning and large-data machine learning.
We derive a theory of human-like few-shot learning from von-Neuman-Landauer's principle.
Modelling human learning is difficult as how people learn varies from one to another.
Under commonly accepted definitions, we prove that all human or animal few-shot learning, and major models
including Free Energy Principle and Bayesian Program Learning that model such learning, approximate our theory, 
under Church-Turing thesis.
We find that deep generative model like variational autoencoder (VAE) can be used to approximate our theory and 
perform significantly better than baseline models including
deep neural networks, for image recognition, low resource language processing, and character recognition.
\blfootnote{Work in progress.}


\end{sciabstract}

\section*{Introduction}
During the past decade, fast progress in deep learning \cite{lecun2015deep} has empowered computer speech recognition, image processing, natural language processing, protein folding, game playing and many other applications. 
However, these great progresses fell short when we try to understand our own learning mechanism: How to model human learning \cite{CS61},
\cite{KF10}, \cite{lake2015human}?

Species in nature learn quickly to survive.
When a dragonfly is hatched, within hours it firms up its wings and then flies to catch mosquitoes; a newborn does not need tons of repeated examples or transfer learning to identify an apple. 
Most human or animal learning exhibits a mixture of inherited intelligence, few-shot learning without prior knowledge, as well as long term many-shot learning. It is interesting to note that these learning programs are encoded in our genomes but they are not all the same, even for individuals within the same species.
The diversity of these learning algorithms is vividly expressed by Spearman's "g" factor~\cite{CS61}. 

Unlike data-laden, model-heavy, and energy-hungry deep learning approaches, most human learning appear to be simple and easy.
Merely scaling up current deep learning approaches may not be sufficient for achieving human level intelligence. We miss certain major
components when modelling human or animal learning. 

Diversity is one of the missing part when modelling human or animal few-shot learning. There are eight billion people on earth, each with a unique few-shot learning model~\cite{stern2017individual}. 
Even if we just want to model one person, a single person often uses
different parameters, features, and perhaps different algorithms to deal with different learning tasks. 
Ideally we want a framework that can cover the diversity in human and animal 
few-shot learning.
Facing such a seemingly formidable task,
traditional thinking in machine learning will only lead us to various traps. To avoid such traps
we need to go back to the very first principles of physics.


Specifically, we start from an agreed-upon law in thermodynamics, to formally derive our model for few-shot learning, and prove this is the optimal model within our 
framework in the sense that all other models including human ones may be viewed as approximations to our framework.
We show a deep connection between our framework and the free energy principle~\cite{KF10} and the
Bayesian Program Learning model~\cite{lake2015human}.
By the end of this process, a component of data compression during the inference
phase of learning emerges as a key component of all few-shot learning models.

First, we formalize our intuitive and commonly accepted concept of human-like few-shot learning. For example,
our definition below is consistent with what is used in \cite{lake2015human}, and in the same
spirit of \cite{KF10}.

\noindent

\begin{definition}
\label{def1}
Consider a universe $\Omega$, partitioned into $H$ disjoint concept classes: ${\cal C}_h$, $h=1,2, \ldots, H$.
Few-shot ($k$-shot) learning is described as follows:
\begin{enumerate}

\item $n$ elements in or outside $\Omega$ are given as unlabelled samples $y_1, \ldots , y_n$;
\item There are $k$ labelled examples for each class ${\cal C}_h$, for small $k$;
\item The learning program, using a computable metric ${\cal M}$, few-shot learns ${\cal C}_h, h=1,2, ... H$,
if it uses the $n$ unlabelled samples and $k$ labelled samples and minimizes the objective function:
\[
\sum_{h=1}^H \sum_{i=1}^{|{\cal C}_h|} {\cal M} (x_i, core_h ) \mid y_1, \ldots, y_n , x_i \in {\cal C}_h,
\]
where $core_h=\psi(k {\rm \ samples \ of \ } \mathcal{C}_h)$ representing a transformed representation of 
the $k$ labelled samples from $\mathcal{C}_h$.
\end{enumerate}
\end{definition}

This definition covers most of our common sense few-shot learning scenarios and other studies. In particular, this is
used in one-shot learning by~\cite{lake2015human}.
As each independent individual, we do not all use a same metric, or even similar metric, to few-shot learning.
For example, MN Hebart et al \cite{HZPB20} identified 49 highly reproducible dimensions to 1854 objects to measure their similarity. 
Different people can be equipped to better observe some of these dimensional features.

We explain the intuition behind~\Cref{def1} via a simple example. A human toddler may have already seen many 
unlabelled samples of fruits which, for example, contains two classes: apples and pears.
Then given a new labelled sample from each class, the toddler learns how to differentiate between these two fruits. The number of labelled data required for one to classify may vary as people have different learning algorithms.

Current deep learning based approaches for few-shot learning generally depend on 1) many auxiliary labelled training samples or task-specific data augmentation for transfer learning or meta learning~\cite{finn2017model};
or 2) very large scale self-supervised pre-training \cite{brown2020language}.
These approaches thus fall short to model few-shot learning in nature by humans and animals as they can hardly account for the diversity in learning algorithms and they either neglect the unsupervised scenario that humans are mostly exposed to or use the scale of unlabelled data and training parameters that are far beyond creatures need.

Many attempts have been made to understand human learning through cognitive, biological, and behavior sciences. 
Some studies have established basic principles a human learning model should obey.
One theory is the two-factor theory of intelligence by Charles Spearman in 1904~\cite{CS61}, where the ``g'' factor is an indicator of the overall cognitive ability, and the ``s'' factor stands for the aptitude that a person possesses in specific areas. As ``g'' factor is genetically-related~\cite{bouchard2009genetic}, it indicates the necessity of a learning theory that can account for the diversity in creatures' learning ability.
Another theory is the Free Energy Principle by Karl Friston \cite{KF10} that
human (and all biological systems) learning tends to minimize the free energy between internal understanding in the sense
of Bayesian (under internal perceived distribution $p$) and that of the environmental event (under distribution $q$), measured by KL-divergence \cite{MB}. 
In a similar spirit, Lake, Salakhutdinov and Tenenbaum~\cite{lake2015human} proposed a 
Bayesian program learning (BPL) model, learning a probabilistic model for each concept and achieve human-level performance.
Two articles by Schmidhuber \cite{JS2009} and by Chater and Vitanyi \cite{ChVi03}
linked simplicity to human cognition and appreciation of arts. 

Instead of exploring a biological basis for few-shot learning, we think it is possible to
mathematically derive an optimal framework that can unify the above theories.
We further demonstrate by experiments that our new model indeed works significantly better than other classical deep learning neural networks for few-shot learning. 
As a byproduct of our new model, a new concept class of "interestingness" is
learned; this class implies where our appreciation of art, music, science and games comes from. Extending this observation, some aspects of
consciousness may be modelled as a set of few-shot learned concepts. Consequently, we hypothesize
the ability of labelling input data becomes a key step to acquiring some aspects of consciousness.

\section*{A theory of few-shot learning}

We mathematically derive an optimal few-shot learning model for \Cref{def1} that is effective and is able to cover enormous diversities existed in different species.
The task may appear to be formidable because of 
conflicting and seemingly very general goals: each individual is allowed to have a different learning model, 
yet our model has just one program to model everybody; we do not yet exactly know the complete
underlying biological mechanisms, yet we need to implement the right functionality; there are infinite number of models, yet we need to 
choose one that is optimal; we are not really interested in "proposing models" out of blue, yet we wish our model to be a 
mathematical consequence of some basic laws of physics;  
the model needs to be theoretically sound, yet practically useful.

For simplicity and readability, we
begin with one-shot learning, $k=1$ in \Cref{def1}. Thus, $core_h$ in \Cref{def1} is just the single labelled sample $x_h$.
For larger $k$, $core_h$ can be some form of average of the $k$ samples.
As \Cref{def1} defined, some unlabelled objects are assumed and it's also possible to extend the definition by adding distribution, learnt from either unlabelled or labelled data, to $\Omega$.
Using metric ${\cal M}$ that is responsible 
for $k$-shot learning of an individual, the learning system seeks to minimize the energy function

\[
\sum_{h=1}^H \sum_{i=1}^{|{\cal C}_h|} {\cal M} (x_i , x_h | y_1, \ldots , y_n),
\]

\noindent
or, assuming ${\cal H } (y_1 , \ldots , y_n )$ is a pre-trained model of $y_1 , \ldots , y_n$,  or other labelled samples, capturing the distribution.

\[
\sum_{h=1}^H \sum_{i=1}^{|{\cal C}_h|} {\cal M} (x_i , x_h | {\cal H} ( y_1, \ldots , y_n )),
\]

Now the question is, what sort of ${\cal M}$ should we use? 
Indeed, this varies from person to person. Can we unify
all such measures, algorithms and inferences? Let's go back to the fundamentals.

\begin{principle}[von-Neuman-Landauer Principle]
    Irreversibly processing 1 bit of information costs 1kT; reversible computation is free.
\end{principle}

Then for two objects $x,y$, the minimum energy needed to convert between $x$ and $y$ in our brain is:
$$
{\cal E}_U(x,y) = \min \{ |p| : U(x,p) = y, U(y, p)= x \},
$$
\noindent
where $U$ is a universal Turing machine or our brain, assuming Church-Turing thesis. Since we can prove a theorem showing all Universal Turing machines are equivalent modulo a constant and efficiency, we will drop the index $U$ (see \cite{LV97}). To interpret, ${\cal E}(x,y)$ is the length of the shortest
program that reversibly converts between $x$ and $y$. These bits used in the shortest program $p$ when they are 
erased will cost $|p|$kT of energy, according to the John von Neuman and Rolf Landuaer's law.
This leads us to a fundamental theorem~\cite{BGLVZ98}:

\begin{figure}[th]
    \centering
    \includegraphics[width=0.8\linewidth]{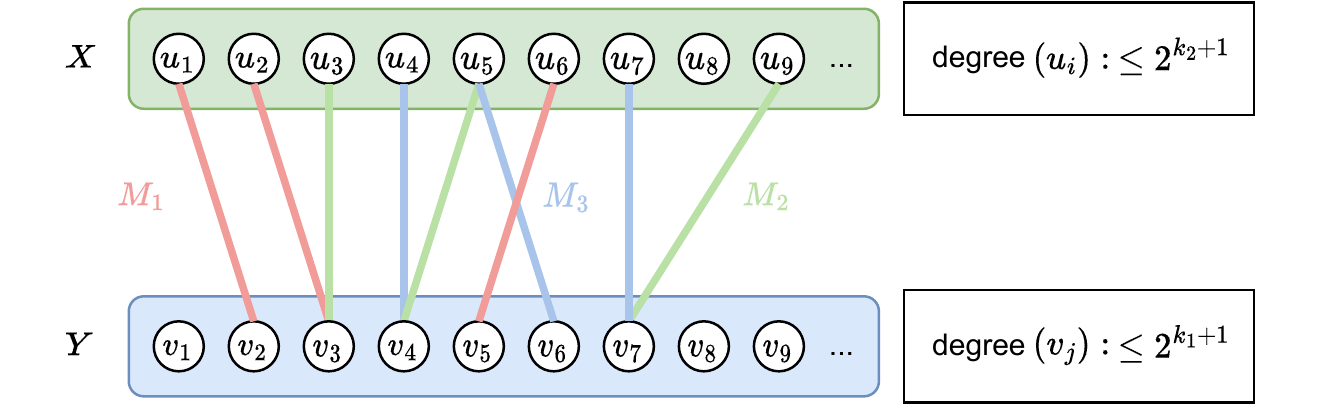}
    \caption{Bipartite Graph}
    \label{fig:bipart}
\end{figure}

\begin{theorem}
\label{thm1}
${\cal E} (x,y) = \max \{ K(x|y), K(y|x) \}+O(1)$.
\end{theorem}
$K(x|y)$ is the Kolmogorov complexity of $x$ given $y$, or informally, the length of the shortest program
that outputs $x$ given input $y$ (details are shown in~\cite{LV97}).
As this theorem was proved thirty years ago and it is
vital in our theory, to help our readers, we will provide an intuitive but less formal proof here.

\begin{proof}

By the definition of ${\cal E}(x,y)$, it follows ${\cal E}(x,y) \geq K(x | y)$ and ${\cal E}(x,y) \geq K(y|x)$, thus we 
have ${\cal E} (x,y) \geq \max \{ K(x|y), K(y|x) \}$.

To prove the other direction ${\cal E} (x,y) \leq \max \{ K(x|y), K(y|x) \}$, 
we need to construct a program \texttt{p} such that \texttt{p} outputs $y$ on input $x$ and \texttt{p} outputs
$x$ on input $y$, and length of \texttt{p} is bounded by $\max \{ K(x|y), K(y|x) \}+O(1)$.

Let $k_1 = K(x|y)$, and $k_2 = K(y|x)$. Without loss of generality, assume $k_1 \leq k_2$.
We first define a bipartite graph $\{ X,Y,E\}$, where $X,Y = \{0,1\}^*$, as shown in~\Cref{fig:bipart} and $E$ is a finite set of
edges defined between $X$ and $Y$ as follows:

\[
E = \{ \{u,v\}, u \in X, v \in Y, K(u|v) \leq k_1, K(v|u) \leq k_2 \}
\]

Note that a particular edge $(x,y)$ is in $E$. If we find edge $(x,y)$, then given $x$, \texttt{p} can output $y$,
and vice versa. So the idea of the proof is to partition $E$ properly so that we can identify $(x,y)$ easily.
Two edges are disjoint if they do not share nodes on either end. 
A matching in graph theory is a set of disjoint edges in $E$.

{\bf Claim.}
$E$ can be partitioned into at most $2^{k_2 + 2}$ matchings.

\textit{Proof of Claim.} Consider edge $(u,v) \in E$. The degree of a node $u \in X$ is bounded by $2^{k_2 +1}$ because
there are at most $2^{k_2 +1}$ different strings $v$ such that 
$K(v|u) \leq k_2$, accumulating possible strings from $i=1$ to $i=k_2$ gives us $\sum_{i=1}^{i=k_2}=2^{k_2+1}-2$. Hence $u$ belongs to at most $2^{k_2 + 1}$ matchings. 
Similarly, node $v \in Y$ belongs to at most $2^{k_1 + 1}$ matchings. We just need to put
edge $(u,v)$ in an unused matching. (End of Proof of Claim)

Let $M_i$ be the matching that contains edge $(x,y)$
We now construct our program \texttt{p}. \texttt{p} operates as follows:

\begin{itemize}
\item
Generate $M_i$ following the proof of Claim, i.e. enumerating the matchings.
This uses information $k_1$, $k_2$, and $i$. $K(i) \leq k_2 + O(1)$
\item
Given $x$, \texttt{p} uses $M_i$ to output $y$, and given $y$, \texttt{p} uses $M_i$ to output $x$.
\end{itemize} 

\end{proof}

A conditional version of Theorem 1, using information in Definition 1, 
can be obtained 
${\cal E}(x,y|y_1, \ldots , y_n ) = \max \{ K(x|y, y_1, \ldots , y_n), K(y | x, y_1, \ldots , y_n) \}$, conditioning on 
unlabelled samples $y_1, \ldots, y_n$.
According to \cite{BGLVZ98}, this distance is universal, in the sense that 
${\cal E} (x,y)$ is the minimum among any other computable distances:

\begin{theorem}
For any computable metric $D$, there is a constant $c$, such that for all $x,y$, ${\cal E} (x,y) \leq D(x,y) + c$.
\end{theorem}

This theorem implies: if $D$ metric finds some similarity between $x$ and $y$, so will ${\cal E}$.
Thus, the above theorem implies, up to some constant $O(H)$
\[
\sum_{h=1}^H \sum_{i=1}^{|{\cal C}_h|} {\cal E} (x_i \in {\cal C}_h, core_h | y_1, \ldots, y_n ) \leq \sum_{h=1}^H \sum_{i=1}^{|{\cal C}_h|} {\cal M} (x_i \in {\cal C}_h, core_h | y_1, \ldots, y_n ).
\]

\noindent
When unlabelled samples $y_1, \ldots , y_n$ plus other irrelevant historical labelled samples are modeled by some 
model ${\cal H}$ such as a generative model (e.g., VAE), then the above inequality can be rewritten as:

\begin{equation}\label{1}
\sum_{h=1}^H \sum_{i=1}^{|{\cal C}_h|} {\cal E} (x_i \in {\cal C}_h, core_h | {\cal H} ) \leq \sum_{h=1}^H \sum_{i=1}^{|{\cal C}_h|} {\cal M} (x_i \in {\cal C}_h, core_h | {\cal H} ).
\end{equation}

Thus, ${\cal E}$ gives optimal metric for few-shot learning algorithm. Other algorithms satisfied \Cref{def1} are the approximation to this optimal solution.
\footnote{Note that ${\cal E}$ is a metric: it is symmetric, and satisfies triangle inequality}

In addition, we show that our theory's deep connection to two well-established principles of learning in neuroscience and psychology.
Friston's Free Energy Principle (FEP)~\cite{KF10}, derived from Bayesian brain hypothesis~\cite{knill2004bayesian}, states that brain seeks to minimize surprises.
Specifically, it assumes the brain has its internal state (a.k.a. generative model) that implicitly models the environment according to the sensory data. Hidden (latent) variables need to be defined for the internal state, which are drawn from prior beliefs. Ideally, these prior knowledge is also modelled, which is made possible by hierarchical generative models. 
The free energy principle (FEP) is often interpreted as Bayesian optimization,
using the Evidence Lower Bound (ELBO) as
$\text{ELBO} = \log p(x;\theta) - D(q(z)\|p(z|x;\theta)$
optimization function.
Here the evidence $\log p(x;\theta)$ is the encoding length of $x$ under probability $p$, and the Kullback-Leibler divergence
term is the p-expected encoding length difference. 
This is half of~\Cref{thm1} and FEP is asymmetric if we view it as a distance. 
However, the symmetry is important to few-shot learning. For example, 
a scarlet king snake may look like a coral snake, but the latter certainly 
has more deadly features the former lacks, one way compression 
$K({\rm Scarlet King Snake} | {\rm Coral Snake})$ is not sufficient to distinguish the two. 
Despite of the fact 
{\it H. influnza} with genome size 1.8 million and {\it E. coli} with genome size 5 million they are sister species but {\it E. coli} 
would be much closer to a 
species with zero genome $G_0$ or just a covid-19 genome with this asymmetric measure ($K( G_0 | E. coli)$
than with {\it H. influnza} ($K(H.~influnza | E.~coli)$). 
A symmetric interpretation of Friston's FEP can be derived by requiring minimum conversion energy as we show in~\Cref{thm1}.

\begin{figure}
    \centering
    \includegraphics[width=0.8\linewidth]{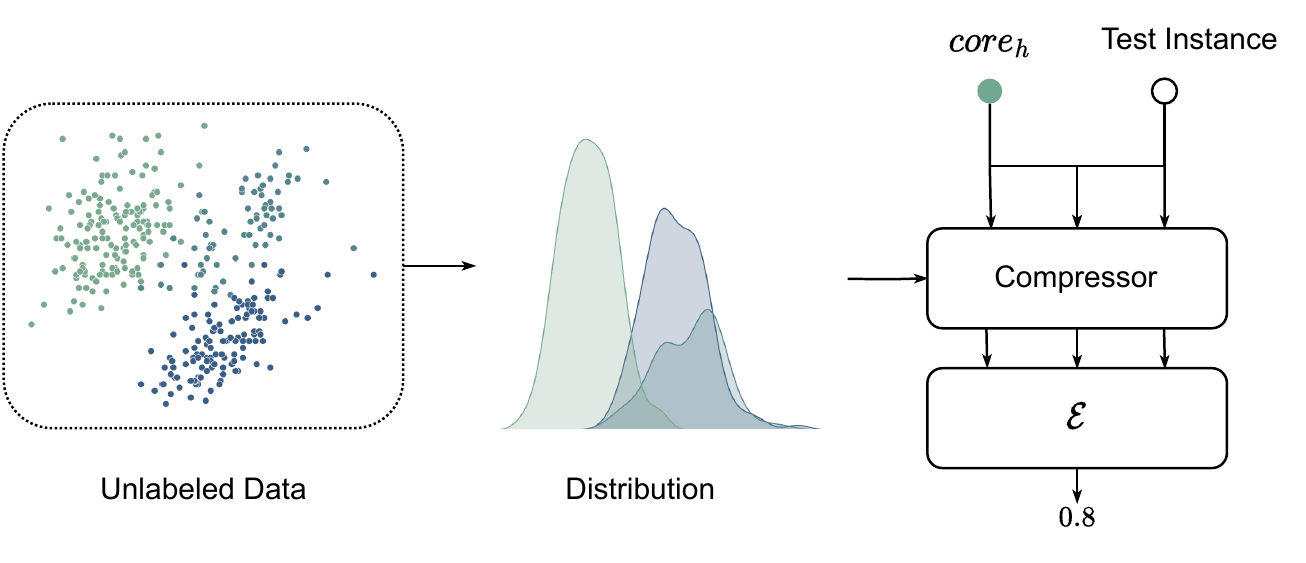}
    \caption{Illustration of our framework, dashed line indicates optional component when learning.}
    \label{fig:arch}
\end{figure}

Different individuals may use different compression algorithms to do data abstraction and inference. 
It can be viewed that these algorithms all approximate $\mathcal{E}(x,y)$. Some are more efficient than others in different situations. The individuals with better compression algorithms have bigger ``g'' factor. Diversified compression algorithms also guarantee better survival chances of a community when facing a pandemic. As compression neural networks are genetically encoded, the ``g'' factor is thus inheritable.
This can be seen via~\Cref{fig:arch}, compression algorithms vary from one to another. The distribution of the data to be learnt is either implicitly or explicitly captured by creatures. Those who can better utilize unlabelled data to capture distribution may have a more efficient compression algorithm.

\section*{Experimental Results}

\subsection*{Image Experiments}

To approximate our universal few-shot learning model, we use a hierarchical VAE as our underlying
model ${\cal H}$ in Inequality \ref{1} to model the unlabelled samples $y_1, \ldots , y_n$. This hierarchical structure coincides with our visual cortex and 
brain structure~\cite{friston2008hierarchical}.
According to integrated information theory~\cite{KT2011}, an input $y$ may come from all sensing terminals: vision, hearing, smell, taste, sensation. 
Often, creatures are exposed to an unsupervised environment where objects are unknown and unlabelled. Revisiting the negative ELBO, we can see it can be interpreted as changing perceptions to minimize discrepancy (minimize KL divergence) or changing observations to maximize evidence, in the context of FEP. When the creatures are exposed to a ``tree'' and they do not fully realize what it is, the sensory information of the objects are internalized with hidden states (inner belief) that can describes how it believes the generation process of a ``tree''. This process of generation, helps the creatures to identify the latent similarities among objects that belong to the same category, without the full awareness. This process of "unconsciously" training to generate helps the creatures to better categorize in future. When the identity of a ``tree'' is finally revealed, they can generalize quickly. This explains our rationale of using a VAE to process unlabelled samples. Consequently, the Kolmogorov complexity 
terms in Inequality \ref{1} are naturally approximated by a VAE based compressor \cite{townsend2018practical}.

To test the hypothesis, we carry out the experiment on five datasets, MNIST, KMNIST, FashionMNIST, STL-10 and CIFAR-10. 
We first train a hierarchical VAE on unlabelled data to learn to generate $\hat{x}$ that's as close to $x$ as possible. This corresponds to the time when creatures exposed to a environment without knowing the object, implicitly learning the latent representation among objects. When the identity of objects are revealed, a VAE based universal compressor can be used to identify the new objects. Specifically, after training a hierarchical VAE unsupervisedly, we compare the ${\cal E}$ energy function between a labelled image and a test image, as in~\Cref{def1}. In our experiment, we use 5 labelled samples per class to test the accuracy of classification. The energy function ${\cal E}$ relies on a compressor to approximate. We thus use the bits-back argument to directly use our trained VAE for the compressor in~\cite{townsend2018practical}. Our result shows that using only 5 samples, our method outperforms traditional 
supervised models like SVM, CNN, VGG and Vision Transformer (ViT) on all five datasets. These supervised methods are chosen to represent different model complexity with wide range of number of parameters. As we can see, when labelled data are scarce, supervised methods are not effective: complex models like VGG cannot perform better than SVM and this tendency is more obvious on ViT without pre-training. The improvement that our method brings is more obvious on more complex datasets like STL-10 and CIFAR-10. Similar result is also obtained in the recent work, across different shot settings~\cite{jiang2022few}.

We also compare with using latent representation directly with $k$-Nearest-Neighbor classifier, labelled as ``Latent'' in the table. The architecture and training procedure for ``Latent'' method is exactly the same to our method --- we train on unlabelled data to generate the sample and then take the latent representation for classification. We can see using latent representation outperforms all supervised methods on four out of five datasets. But the accuracy is still way lower than our method, indicating our method can better utilize the generative models.

\begin{table}[]
    \centering
    \begin{tabular}{c|c|c|c|c|c}
    \toprule
         & MNIST & KMNIST & FashionMNIST & STL-10 & CIFAR-10  \\
         \hline
        SVM & 69.4$\pm$2.2 & 40.3$\pm$3.6 & 67.1$\pm$2.1 & 21.3$\pm$2.8 & 21.1$\pm$1.9 \\
        CNN & 72.4$\pm$3.5 & 41.2$\pm$1.9 & 67.4$\pm$1.9 & 24.8$\pm$1.5 & 23.4$\pm$2.9 \\
        VGG & 69.4$\pm$5.7 & 36.4$\pm$4.7 & 62.8$\pm$4.1 & 20.6$\pm$2.0 & 22.2$\pm$1.6 \\
        ViT (disc) & 58.8$\pm$4.6 & 35.8$\pm$4.1 & 61.5$\pm$2.2 & 24.2$\pm$2.5 & 22.3$\pm$1.8 \\
        Latent & 73.6$\pm$3.1 & 48.1$\pm$3.3 & 69.5$\pm$3.5 & 31.5$\pm$3.7 & 22.2$\pm$1.6 \\
        Ours & 77.6$\pm$0.4 & 55.4$\pm$4.3 & 74.1$\pm$3.2 & 39.6$\pm$3.1 & 35.3$\pm$2.9 \\
    \bottomrule
    \end{tabular}
    \caption{5-shot image classification accuracy on five datasets.}
    \label{tab:my_label}
\end{table}

\subsection*{Text Experiments}
Our theory is generally applicable, even without pre-training on unlabelled data. Here, we demonstrate significant advantages of our approach with a simple compressor \textit{gzip} over lower resource languages.

\paragraph{Languages with Abundant Resources}
We first test our method on datasets with abundant resources.
Specifically, we compare with three datasets --- AG News, SogouNews and DBpedia.
Similar to image classification, we compare with both supervised methods, including fasttext~\cite{joulin2017bag}, BiLSTM~\cite{schuster1997bidirectional} with attention mechanism~\cite{wang2016attention} and Hierarchical Attention Network (HAN)~\cite{yang2016hierarchical}, and non-parametric methods that use Word2Vec (W2V)~\cite{mikolov2013efficient} as representation. We also compare with pre-trained language models like BERT~\cite{devlin2019bert} We use five labelled data for each class (5-shot) for all the methods.

\begin{table}[t]
    \centering
    \begin{tabular}{c|c|c|c}
    \toprule
         & AG News & SogouNews & DBpedia  \\
         \hline
        fasttext & 27.3$\pm$2.1 & 54.5$\pm$5.3 & 47.5$\pm$4.1 \\
        Bi-LSTM+Attn & 26.9$\pm$2.2 & 53.4$\pm$4.2 & 50.6$\pm$4.1 \\
        HAN & 27.4$\pm$2.4 & 42.5$\pm$7.2 & 35.0$\pm$ 1.2 \\
        W2V & 38.8$\pm$18.6 & 14.4$\pm$0.5 & 32.5$\pm$11.3 \\
        BERT & 80.3$\pm$2.6 & 22.1$\pm$4.1 & 96.4$\pm$4.1 \\
        Ours & 58.7$\pm$4.8 & 64.9$\pm$6.1 & 62.2$\pm$2.2 \\
    \bottomrule
    \end{tabular}
    \caption{5-shot text classification accuracy on three datasets.}
    \label{tab:tc_id}
\end{table}
Surprisingly, even without any pre-training and with a simple compressor like \textit{gzip}, our method outperforms all non-pretrained supervised methods and non-parametric methods in low data regime. This indicates that compressor serves as an efficient method to capture the regularity and our information distance is effective in comparing the similarity based on the essential information. When comparing with pre-trained models like BERT, we can see our method is significantly higher on SogouNews, a special dataset that includes Pinyin --- a phonetic romanization of Chinese, which can be viewed as an Out-Of-Distributed (OOD) dataset as it uses the same alphabet as \texttt{english} corpus.

\begin{table}[t]
    \centering
    \begin{tabular}{c|c|c|c|c}
    \toprule
       & Kinnews & Kirnews & Swahili & Filipino \\
       \hline
       BERT  & 24.0$\pm$6.0 & 38.6$\pm$10.0 & 39.6$\pm$9.6 & 40.9$\pm$5.8 \\
       mBERT & 22.9$\pm$6.6 & 32.4$\pm$7.1 & 55.8$\pm$16.9 & 46.5$\pm$4.8\\
       Ours & 45.8$\pm$6.5 & 54.1$\pm$5.6 & 62.7$\pm$7.2 & 65.2$\pm$4.8\\
    \bottomrule
    \end{tabular}
    \caption{5-shot text classification accuracy on low-resource datasets}
    \label{tab:tc_ood}
\end{table}

\paragraph{Low-Resource Languages}
Sufficiently pre-trained language models are exceptional few-shot learners~\cite{brown2020language}.
However, when faced with low resource data or distributions that are significantly different from any pre-trained data, those pre-trained language models lose their advantages to our method.
We compare our method with BERT on
four different low-resource language datasets - \texttt{Kinyarwanda}, \texttt{Kirundi}, \texttt{Swahili} and \texttt{Filipino}. These datasets are curated to have the Latin alphabets, same as \texttt{english} corpus. BERT has performed extremely well as shown in~\Cref{tab:tc_id} due to pre-training on billions of tokens. However, when facing low-resource datasets, BERT perform significantly worse than our method only using \textit{gzip} as we can see in~\Cref{tab:tc_ood}, no matter using multilingual pre-trained version or the original one. Note that mBERT is pre-trained on 104 languages including \texttt{Swahili} and \texttt{Tagalog} (on which Filipino is based on). As we can see on \texttt{Swahili} and \texttt{Filipino}, mBERT performs better than BERT, but still significantly lower than our method.

\subsection*{Omniglot one-shot-classification dataset}
\begin{wrapfigure}{r}{0.5\textwidth}
  \begin{center}
    \includegraphics[width=0.48\textwidth]{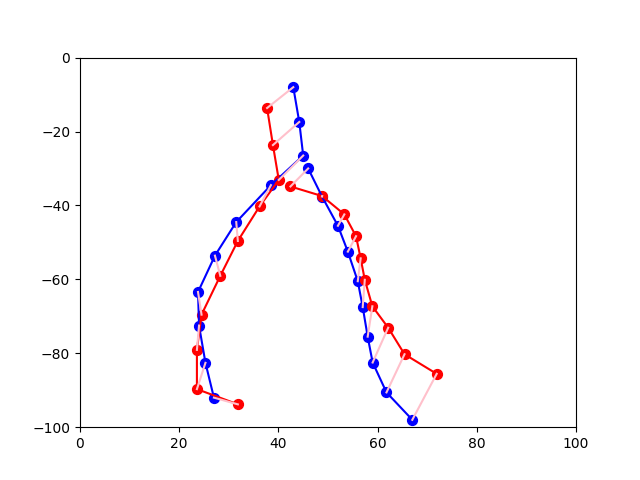}
  \end{center}
    \caption{Distance between two Bezier curves}
    \label{fig:Bezier}
\vspace{-1em}
\end{wrapfigure}

In \cite{lake2015human}, a one-shot learning framework 
Bayesian program learning (BPL) was proposed. It learns a simple probabilistic model for each concept. Taking a negative logarithm converts a Bayesian formula to a description length paradigm, hence BPL can be viewed as one particular approximation to our theory. 
Here we provide another simple approximation of our theory for the Omniglot one-shot-classification dataset of \cite{lake2015human}. 

Our system first decompose a given character into strokes, then compute ${\cal E}(a,b)$ between characters $a$ and $b$,
using all their possible stroke decomposition.
We provide how to calculate ${\cal E}(a,b)$ here and details of decomposition program is given in~\Cref{appx:a}.

\begin{enumerate}
\item
Fit a stroke by a Bezier curve;
\item
Ensure the number of points on two curves are same. This algorithm utilize equally split method to select certain same number of points on each curve Figure~\ref{fig:Bezier};
\item
Ensure the area of the convex hull and the barycenter of the compared 
characters are the same;
\item
Use max Cartesian distance between parallel points on two Bezier curves to 
approximate the minimum encoding distance
between two Bezier curves, as shown in Figure~\ref{fig:Bezier};
\item
Choose the character with minimum distance.
\end{enumerate}

This simple implementation achieves $92.25\%$ accuracy 20-way-1-shot on this dataset. 
The point here is to demonstrate various approximations of our theory that work rather 
than comparing accuracy. 
At $96.75\%$~\cite{lake2015human} or at $92.25\%$ might be two different individuals 
with different compression algorithms.





\section*{Unification} 

Our framework can unify other popular deep neural networks for few-shot learning.

\textbf{Siamese Network:} Siamese network uses twin subnetwork to rank the similarity between two inputs in order to learn useful features. $\mathcal{M}$ here is often a contrastive loss. This framework shows strong performance in one-shot image recognition~\cite{koch2015siamese}.

\textbf{Prototypical Network:} Prototypical networks~\cite{snell2017prototypical} propose to optimize the distance metric $\mathcal{M}$ directly by learning $core_h$ in representation space. $core_h$ are represented as the mean of embedded support samples.

\textbf{Bi-Encoder:} In the context of natural language processing, one of the dominant structure is the Bi-Encoder design with each encoder being a pre-trained language model. For example, in information retrieval, Dense Passage Retrieval (DPR), with two encoders encoding query and document respectively, has become the new state of the art. To capture semantic similarity, sentenceBERT~\cite{reimers2019sentence} also adopts the bi-encoder design and becoming one of the most prevalent methods for semantic textual similarity. $\mathcal{M}$ in both cases can either be cosine similarity or Euclidean distance between the representation learned through pre-trained models.

\textbf{Information Distance based Methods:}
Hundreds of algorithms were published, before the deep learning era, on parameter-free data mining, clustering, anomaly detection, classification
using information distance $\mathcal{E}$~\cite{LBCKKZ01, keogh2004towards, bennett2003chain, nykter2008critical, benedetto2002language, nykter2008gene}, with a comprehensive list in \cite{LV97}. Recently \cite{jiang2022few} have discovered using information distance with deep neural networks and leverage the generalizability of few-shot image classification.
%
This work shows that with the help of deep generative models, unlabelled data can be better utilized for few-shot learning under our framework.

\section*{Conclusion and a discussion on consciousness}

We have defined human-like few-shot learning and derived an optimal form of such few-shot learning. 
Note there is an interesting difference between our theory and classical learning theory.
In classical learning theory, it is well-known that if we compress training data to a smaller 
consistent description, whether it is a classical Bayesian network or a deep neural 
networks~\cite{bengio2009learning, LV97}, we would achieve learning. In this paper, 
we demonstrate that in the inference stage, compression is also important, especially when there are not enough labelled data to train a small model. On the biological
side, compression circuits using 
predictive coding in human cortex has been studied by~\cite{Rao}.
Experiments have also strongly supported our theory. We expect to see more practical systems approximating
our theory can be implemented to solve commonplace few-shot learning problems when large amounts of labelled data for deep learning
is lacking. We now wish to explore two consequences of our few-shot learning model, to consciousness.

\subsection*{A binary classifier of interestingness}\label{interesting}
Our few-shot learning model has a by-product. We have proved compression is a universal goal that
few-shot learning algorithms approximate. 
Thus this implies immediately a (subconscious) {\it binary classifier}: if something is compressed, then something 
interesting happens, and attention is given. It turns out that this "Interestingness" has been theoretically studied as logical depth
first proposed by Charles Bennett~\cite{LV97}. According to Bennett, a structure is deep if 
it is superficially random but subtly redundant. When few-shot learning happens, significant compression happens, and these deep objects gain attention.
Such a binary classifier might explain our appreciation of
arts, music, games, and science, since these all share a common feature of dealing with non-trivially compressible
objects: whether it is a shorter description of the
data that gives rise of Newton's laws~\cite{LV97}, or a piece of art or music that itself is compressible or that reminds us of something we have experienced before, hence very compressible, we feel we understand it and hence appreciate it. 
Science is nothing but compressing data into simpler descriptions of nature. 

\subsection*{Consciousness and the ability of labelling data}


Do other species have consciousness? It is difficult to answer this question as consciousness is not testable. 
Thomas Nagel \cite{N74} made a comment: We will never know if a bat is conscious because we are not bats. 

Consider an alternative data-driven approach by asking what a
species can do instead of how they feel. That is, if we treat some aspects of consciousness as a collection of learned concepts, 
then given a compression network, 
the ability of acquiring the relevant concepts becomes a matter of labelling relevant data.
We know learning and consciousness are both located at posterior cortex region \cite{koch2018}. This is in agreement with some injured patients when they lost consciousness.
This is also in agreement with ``bistable perception'' training results with monkeys \cite{GM05}. 

Varieties of consciousness are being pragmatically studied~\cite{Birch2020}. 
These include:
1) the ability of consciously perceive the environment;
2) the ability of evaluating conscious emotions;
3) the ability of having a unified conscious experience;
4) the ability of integrating across time as a continuous stream, one moment flowing into the next; 
5) the conscious awareness of oneself as distinct from the world outside. Many of these abilities may be seen as a 
few-shot learnable concepts, given properly labelled data.

Different animals have various levels of some of such 
consciousness by passing certain tests. For example, chimpanzees, dolphins, Asian elephants, and magpies can recognize themselves by
passing some mirror-mark tests. The corvids display some emotions, and are able to plan ahead. Octopus have powerful perceptual 
facilities obtaining and processing data independently with each tentacle.
Experimentally, awareness emerges when information travels back and forth between brain areas \cite{Boly2011} 
instead of a linear chain of command. 

According to our theory, the brain really only needs to use a universal compressor to compress information, regardless of one processor 
in the head or a few processors in the tentacle (in case of Cephalopods). Thus we can conjecture that 
"consciousness'' then is a matter of 
{\it ability of labelling the data} from sensory terminals. 
Food or enemy in the environment are easy to label. Emotional labelling requires some level of
abstraction. Self-awareness of ``me'' and ``others'' thus is just another binary
classifier trainable depending on if the species is able to do ``displaced reference'' mental labelling.
Other than the human beings, only orangutans are known to have limited displaced reference ability~\cite{Lyn2014}. 

Thus we have just reduced the non-testable question of whether an animal has consciousness in some aspects to if it is able to 
label the corresponding data properly. 

\section*{Acknowledgement}
We thank Dr. Hang Li for suggestions and bringing \cite{MTS2022} to our attention and Dr. Amy Sun for bringing
\cite{scherr2020one} to our attention. The work is supported in part by
Canada's NSERC operating grant OGP0046506, Canada Research Chair Program, and the
Leading Innovative and Entrepreneur teams program of Zhejiang, number 2019R02002, and NSFC grant 61832019.

%
%

\bibliography{papers}
\bibliographystyle{Science}

\appendix
\section{Algorithm for extracting strokes from a character}
\label{appx:a}

Repeat until all pixels of a character are marked, by depth-first search: \\

(1) Extract its skeleton so that the stroke width is 1 pixel point. Then convert
the image to a graph and shrink adjacent cross points.
(2) Randomly select an endpoint as starting point, endpoint at top left has
a greater chance of being selected. Walk until a cross point or endpoint. 
If there is a circle then select a cross point of a top left point if there 
is no cross point. Record this stroke and mark it on the character. 
Allow small number of marked pixel points to make the decomposition more natural.
(3) When meeting a cross point, then enumerate two situations of pen-up and turning, randomly. 
Pen-up means end of a stroke, go to step (2) with the marked graph. Turning
means continuation hence repeat step (2). 
If walking to an endpoint, then attempt to turn by going back to find a new unmarked pixels within 
some small number of pixels or directly end the stroke and repeat step (2) with marked graph.


\end{document}